\documentclass[lettersize,journal]{IEEEtran}
\usepackage{etoolbox}
\makeatletter
\patchcmd{\@makecaption}
  {\scshape}
  {}
  {}
  {}
\makeatother
\usepackage{amsmath,amsfonts}
\usepackage{algorithmic}
\usepackage{algorithm}
\usepackage{array}
\usepackage[caption=false,font=normalsize,labelfont=sf,textfont=sf]{subfig}
\usepackage{textcomp}
\usepackage{stfloats}
\usepackage{booktabs}
\usepackage{bm}
\usepackage{url}
\usepackage{hyperref}
\usepackage[dvipsnames]{xcolor}
\usepackage{makecell}
\usepackage{verbatim}
\usepackage{multirow}
\usepackage{graphicx}
\usepackage{cite}
\begin{document}

\title{Peer is Your Pillar: A Data-unbalanced Conditional GANs for Few-shot Image Generation}

\author{$\text{Ziqiang Li}^{1} \quad \text{Chaoyue Wang}^{2} \quad \text{Xue Rui}^{1} \quad \text{Chao Xue}^3 \quad \text{Jiaxu Leng}^4 \quad \text{Bin Li}^{1,5} $ \\
$^1 \text{Big Data and Decision Lab, University of Science and Technology of China.}$ \\
 \quad $^2 \text{The University of Sydney}$ 
 \quad $^3 \text{JD Explore Academic}$ \quad $^4 \text{School of computer science, Chongqing University of Posts and Telecommunications}$\\
$^5 \text{CAS Key Laboratory of Technology in Geo-spatial Information Processing and Application System.}$\\
{\tt\small iceli@mail.ustc.edu.cn, chaoyue.wang@outlook.com, ruixue27@mail.ustc.edu.cn} \\ {\tt\small 18511867420@163.com, lengjx@cqupt.edu.cn, binli@ustc.edu.cn}
\thanks{Corresponding author: Chaoyue Wang and Bin Li.}
}

\markboth{Journal of \LaTeX\ Class Files,~Vol.~14, No.~8, August~2021}%
{Shell \MakeLowercase{\textit{et al.}}: A Sample Article Using IEEEtran.cls for IEEE Journals}


\maketitle

\begin{abstract}
Few-shot image generation aims to train generative models using a small number of training images. When there are few images available for training (e.g. 10 images), Learning From Scratch (LFS) methods often generate images that closely resemble the training data while Transfer Learning (TL) methods try to improve performance by leveraging prior knowledge from GANs pre-trained on large-scale datasets. However, current TL methods may not allow for sufficient control over the degree of knowledge preservation from the source model, making them unsuitable for setups where the source and target domains are not closely related. To address this, we propose a novel pipeline called Peer is your Pillar (PIP), which combines a target few-shot dataset with a peer dataset to create a data-unbalanced conditional generation. Our approach includes a class embedding method that separates the class space from the latent space, and we use a direction loss based on pre-trained CLIP to improve image diversity. Experiments on various few-shot datasets demonstrate the advancement of the proposed PIP, especially reduces the training requirements of few-shot image generation.
\end{abstract}


\section{Introduction}
\label{sec:intro}

\IEEEPARstart{G}{enerative}  Adversarial Networks (GANs) \cite{goodfellow2014generative} have achieved remarkable performance in generating high-fidelity and diverse synthetic images \cite{zhou2023bc,wang2023clip2gan}, but rely heavily on large-scale, single-class datasets. For instance, StyleGAN2 \cite{karras2020analyzing} is trained on a dataset of 70K high-resolution facial images. However, when the available training data is limited \cite{zhou2022meta,li2023knowledge,dang2023counterfactual}, GANs tend to suffer from issues such as discriminator overfitting \cite{yang2021data,karras2020training} and latent discontinuity \cite{li2022fakeclr,kong2021smoothing}, which can significantly damage the quality and diversity of the generated images. For few-shot image generation (FSIG), we expect three desired properties: (i) \textit{High fidelity.} The generated images should be photo-realistic. (ii) \textit{Enough diversity.} The generator should not simply replicate the training data. (iii) \textit{Target-domain consistency.} Synthetic images should have coherent content and style with the few-shot training data.

\begin{figure*}[t!]
\setlength{\abovecaptionskip}{0.1cm}
    \setlength{\belowcaptionskip}{-0.3cm}
	\centering
	\includegraphics[scale=0.75]{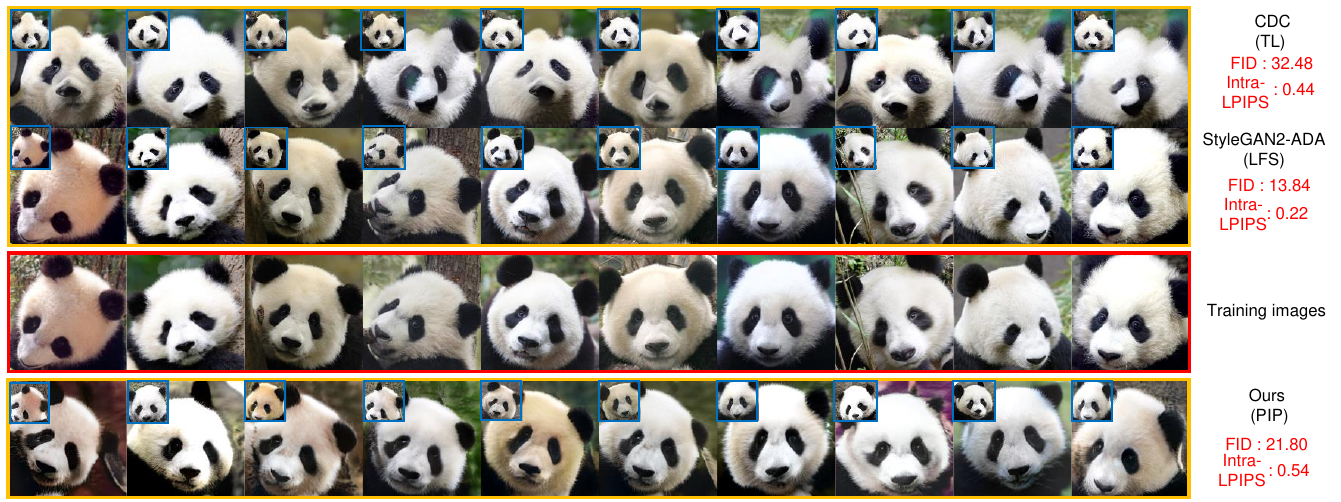}
	\caption{The \textit{Middle} part shows the \textcolor[RGB]{255,0,0}{reference image} from the 10-shot training dataset with red box. The \textit{Top} part displays the generated images and corresponding metrics (FID \cite{heusel2017gans} and Intra-LPIPS \cite{ojha2021few}) of the TL baseline CDC \cite{ojha2021few} and LFS baseline StyleGAN2-ADA \cite{karras2020training}. The large images are generated images that are \textcolor[RGB]{255,192,0}{Down similar} to the reference image on training images, and the small images are generated images that are \textcolor[RGB]{0,112,192}{Top similar} to the reference image on training images. To generate these images, we first generated 4,000 images and assigned each of them to one of the 10 target samples, forming 10 clusters. We then selected the cluster with the reference image and computed the LPIPS distance between the images within the reference cluster and the reference image. The \textcolor[RGB]{0,112,192}{Top similar} and \textcolor[RGB]{255,192,0}{Down similar} images are the images with the lowest and highest LPIPS distance, respectively. Note that the Panda dataset only contains 100 images, so the large synthesis diversity will cause a significant increase in the FID. Therefore, replicating the training images on StyleGAN2-ADA produces the best FID results, but the diversity measured by Intras-LPIPS is poor. The \textit{Bottom} part shows the generated images of our method, which exhibit various fidelity levels different from reference images and achieve competitive FID and Intra-LPIPS metrics.}
	\label{FIG:motivation}
\end{figure*}

To address these challenges, previous studies have explored various data augmentation \cite{zhao2020differentiable,zhao2020image,karras2020training,tran2020towards,jiang2021deceive,li2023exploring} and regularization \cite{yang2021data,li2022fakeclr,tseng2021regularizing,li2023systematic,9851855,chen2023mitigating} techniques for training GANs from scratch, known as the \textbf{Learning From Scratch (LFS)} approach. Although LFS can produce images with \textit{high fidelity} and \textit{target-domain consistency}, it often lacks \textit{enough diversity} in the generated images, resulting in the replication of the training data. To address this issue, some previous studies \cite{li2023styo,nitzan2023domain,wu2023domain,ojha2021few,robb2020few,xiao2022few,zhao2022closer,zhang2022towards,yang2021one,NEURIPS2020_b6d767d2,zhang2022generalized} have explored \textbf{Transfer Learning (TL)}, which leverages the prior knowledge of GANs pre-trained on large-scale datasets and adapts it to a small target domain dataset. While TL can achieve high fidelity, enough diversity, and target-domain consistency, it requires a semantically related large source domain dataset to pre-train the model, which can be challenging to obtain in some domains. Additionally, TL's creativity is limited without access to the source data, and it can only accomplish few-shot image generation through extensive inheritance from the source dataset. For instance, the popular TL method, CDC \cite{ojha2021few}, cannot adapt the AFHQ-CAT source generator to the 10-shot panda dataset, as shown in Fig. \ref{FIG:motivation}. Overall, TL struggles to transfer domain-independent knowledge to the target dataset, which is critical for few-shot image generation.

We propose a new pipeline called Peer is your Pillar (PIP) for few-shot image generation that eases the requirements of the TL method for semantic relevance between source and target datasets and a large amount of source dataset. PIP combines the target few-shot dataset with a peer dataset to create a data-unbalanced conditional generation task, allowing domain-independent knowledge to be shared across the two datasets through collaborative training. This approach offers several advantages: (i) \textit{Limited data dependence.} Unlike TL \cite{li2023styo,nitzan2023domain,wu2023domain}, which requires pre-training models on the source domain with tens of thousands of images, PIP can work with peer datasets that typically contain 1K or even 0.5K images. (ii) \textit{Relaxed semantic similarity.} PIP has a more relaxed semantic similarity requirement compared to TL, making it more practical. As shown in Fig. \ref{FIG:motivation}, PIP is the only method that can achieve good performance in all three properties on the 10-shot Panda dataset.


The straightforward way to exploit conditional GANs in PIP will introduce overly strong category conditions, which results in limited knowledge gained from large-scale peer datasets. To address it, we propose a novel class embedding method that separates the class space from the latent space. By reducing the intensity of class injection, we can leverage more knowledge from large-scale peer datasets, resulting in improved diversity for target few-shot generation. Additionally, to ensure sufficient diversity, we use the direction loss \cite{gal2021stylegan} realized by pre-trained Contrastive-Language-Image-Pretraining (CLIP) \cite{radford2021learning}, which guarantees that the image-pair direction is equal to the CLIP-space text direction between the peer and target domains. Our proposed method has been validated in terms of fidelity, diversity, and target-domain consistency across multiple 10-shot generation datasets.

In summary, our contributions are as follows:

\begin{itemize}
\setlength{\itemsep}{0pt}
\setlength{\parsep}{1pt}
\setlength{\parskip}{1pt}
\item We propose a novel pipeline, \textbf{PIP}, that learns knowledge from peer datasets for few-shot image generation.
\item We design a class embedding method that provides soft condition injection and use the direction loss realized by CLIP to improve the diversity of generated images significantly.
\item We validate the performance of our method across multiple few-shot generation datasets, and our approach is the first to achieve diversity generation on the 10-shot Panda.
\end{itemize}


\section{Related Work}
\subsection{Generative Adversarial Networks}
Generative Adversarial Networks (GANs) \cite{goodfellow2014generative} aim to learn the underlying real distribution by training a generator and a discriminator through a min-max game. However, GANs are notorious for their instability during training, and several approaches have been proposed to mitigate this issue, including the use of efficient architectures \cite{karras2017progressive,brock2018large,karras2020analyzing}, loss functions \cite{gulrajani2017improved,wang2019evolutionary}, optimization methods \cite{heusel2017gans,brock2018large}, regularization \cite{yang2021data,mescheder2017numerics,10.1145/3569928}, and normalization \cite{miyato2018spectral}. Recently, Style-based generators \cite{karras2019style,karras2020analyzing,karras2021alias,sauer2022stylegan} have shown promising results, even on high-resolution and diverse datasets such as ImageNet \cite{sauer2022stylegan}. Another approach is to introduce class information into GANs through class embedding, which enables conditional generation (cGANs) \cite{brock2018large,karras2020analyzing,sauer2022stylegan}. However, the layer-wise structure used by both BigGAN \cite{brock2018large} and StyleGANs \cite{karras2020analyzing,sauer2022stylegan} involves too much conditional injection, which limits domain-independent knowledge sharing.
\subsection{Data-efficient GANs}
While Generative Adversarial Networks (GANs) have achieved remarkable performance in image generation with sufficient training data, in some practical applications, the available training data is limited. To address this, recent research has focused on developing Data-efficient GANs (DE-GANs) \cite{li2022comprehensive}. DE-GANs aim to fit the entire data distribution using limited training samples \cite{yang2021data,tseng2021regularizing,karras2020training,li2022fakeclr,ojha2021few}. Based on the scale of the training dataset, DE-GANs can be divided into two tasks: Limited-data Image Generation and Few-shot Image Generation. While both tasks have similar settings and motivations, the main challenges and employed techniques differ.

\subsubsection{Limited-data Image Generation}
Limited-data image generation refers to GANs that are specifically designed to generate high-quality images with limited amounts of training data, typically 1K or 5K samples. One of the major challenges in this area is discriminator overfitting, which can lead to unstable training and unreasonable decision boundaries. To address this issue, various techniques have been developed, such as data augmentation and regularization. Data augmentation is a popular approach to mitigate overfitting, and several methods \cite{zhao2020differentiable,karras2020training,tran2020towards,jiang2021deceive,zhao2020image} have been proposed for GANs training, including Differentiable Augmentation (DA) \cite{zhao2020differentiable}, Adaptive Discriminator Augmentation (ADA) \cite{karras2020training},  Adaptive Pseudo Augmentation (APA) \cite{jiang2021deceive}. Among these, ADA \cite{karras2020training} has gained impressive performance and is widely used in most DE-GANs due to its adaptive strategy for controlling the strength of augmentations.
Regularization \cite{yang2021data,li2022fakeclr,tseng2021regularizing} is another effective solution to mitigate discriminator overfitting by introducing prior and additional tasks.  Among them, InsGen \cite{yang2021data} and FakeCLR \cite{li2022fakeclr} add contrastive terms to Limited-data generation, stabilizing the training and achieving impressive results.

\subsubsection{Few-shot Image Generation (FSIG)}

Compared to limited-data image generation, few-shot image generation (FSIG) employs significantly less training data, sometimes as little as one-shot, 10-shot, or 100-shot. While learning from scratch (LFS) methods proposed in limited-data GANs can mitigate discriminator overfitting and stabilize GAN training in FSIG, generator collapse tends to replicate training samples, presenting a major challenge for few-shot image generation. To alleviate generator collapse, Kong \textit{et al.} \cite{kong2021smoothing} introduce some regularization technologies for few-shot GANs, which partially relieve the overfitting of LFS methods, but the generated images' quality, diversity, and consistency with the target domain remain limited.  Furthermore, most studies \cite{mo2020freeze,ojha2021few,robb2020few,xiao2022few,zhao2022closer,yang2021one,NEURIPS2020_b6d767d2,zhao2022few,zhao2022few} rely on transfer learning (TL) to leverage knowledge from large-scale source datasets. Some of these studies ensure cross-domain consistency between the source and target domains through a consistency loss based on Kullback-Leibler (KL) divergence \cite{ojha2021few,xiao2022few} or contrastive term \cite{zhao2022closer} to improve few-shot generation diversity. Additionally, some studies \cite{gal2021stylegan,zhang2022towards,zhu2021mind} leverage the semantic power of large contrastive-language-image-pretraining (CLIP) \cite{radford2021learning} models to define the domain-gap direction in CLIP embedding space, which guides generator optimization. Finally, to leverage prior knowledge from the source model, some studies have frozen lower layers \cite{mo2020freeze}, important parameters \cite{robb2020few,NEURIPS2020_b6d767d2,zhao2022few}, or even the entire set of parameters of the source model \cite{yang2021one,alanov2022hyperdomainnet,kim2022dynagan}. Despite impressive synthesis quality, TL methods are often limited in practice, as they require large-scale source domain datasets that are semantically related to the target domain. Therefore, they perform poorly on setups where the source and target domains are not closely related.

\begin{figure*}[t!]
\setlength{\abovecaptionskip}{0.1cm}
    \setlength{\belowcaptionskip}{-0.3cm}
	\centering
	\includegraphics[scale=0.47]{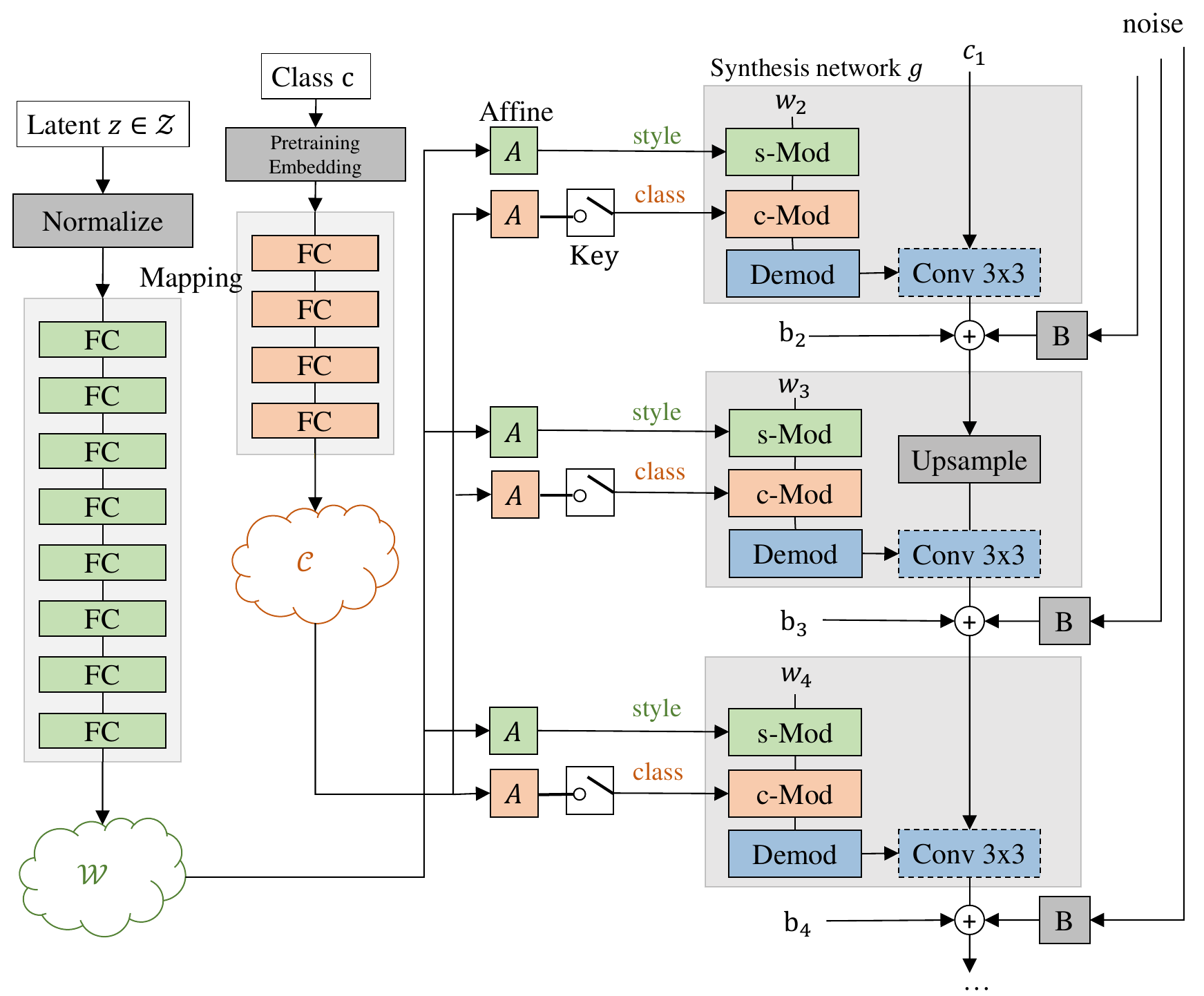}
	\caption{The architecture of our method. Our architecture differs from the baseline StyleGAN2 \cite{karras2020training} architecture in that we incorporate an additional lightweight class mapping network to build the class latent space ($\mathcal{C}$), instead of blending the latent code ($z$) and class code ($c$) before a common mapping network. Additionally, we adopt another affine module to acquire the class vector, which adjusts the synthesis network through class modulation (c-Mod). To improve the diversity of few-shot generation and share knowledge between different classes, we use pre-trained class embeddings and the Key module to control the strength of the class modulation.}
	\label{FIG:structure}
\end{figure*}
\section{Methodology}
The objective of this paper is to generate images with high fidelity, adequate diversity, and target-domain consistency using only a small amount of training data. Previous pipelines, such as TL, require large source datasets that are semantically related to complete the generation task, making it unsuitable for tasks where the source and target domains are not closely related. To overcome this limitation, we propose a new pipeline called Peer is Your Pillar (PIP) in Sec. \ref{section_3.1}. To further enhance the realism and diversity of the generated images, we introduce class embedding and modulation in Sec. \ref{section_3.2} and regularization in Sec. \ref{section_3.3}. Our baseline model is StyleGAN2-ADA \cite{karras2020training}.


\subsection{The Pipeline of Peer is your Pillar (PIP)}
\label{section_3.1}

Training Generative Adversarial Networks (GANs) can be challenging, especially when training data is limited. Discriminator overfitting and generator collapse are common issues in this scenario, particularly for conditional generation \cite{karras2020training, shahbazi2022collapse}. One solution for few-shot image generation is to acquire and reuse knowledge from large-scale datasets. However, Transfer Learning (TL) has limited creativity since it only offers a pre-trained model without access to the source data. TL's performance can suffer when working on datasets without semantically related large source datasets since too much target domain knowledge needs to be created beyond the source domain. To address this issue, we propose Peer is your Pillar (PIP), which uses co-training to gain knowledge from peer datasets to improve performance on target datasets. PIP trains conditional GANs on a data-unbalanced dataset containing a large-scale peer dataset and few-shot target datasets to cover the few-shot image generation task. Our approach shares more knowledge from the peer dataset to the target dataset, leading to the first diversity generation on tasks where the peer and target domains are not closely related. Notably, we found that conditional GANs on data-unbalanced datasets have not met model collapse, which was observed in data-balanced conditional generation tasks \cite{shahbazi2022collapse}. However, directly adopting the conditional StyleGAN2-ADA \cite{karras2020training} has limited diversity and reality. Therefore, we introduce gradual changes to the baseline to improve the performance of PIP.

\subsection{Architecture}
\label{section_3.2}

PIP aims to improve the diversity of few-shot image generation by sharing knowledge from the peer dataset through data-unbalanced conditional GANs. The form of class embedding plays a crucial role in controlling the shared knowledge. In the original StyleGAN2 \cite{karras2020analyzing}, the latent code ($z$) and class code ($c$) are blended before a common mapping network, which provides too much class information and limits the information sharing from the peer dataset. To address this issue, we introduce two modifications in this section, namely, \textbf{Pretrained Class Embedding} and \textbf{Separable Class Modulation}, as illustrated in Fig. \ref{FIG:structure}.

\noindent\textbf{Pretrained Class Embedding.} The original conditional GANs use the one-hot class label to control the sample class and improve overall performance. However, this approach creates hard labels, and the distance between different classes is the same. To soften the label, we adopt a pre-trained VGG-19 \cite{simonyan2014very} network (Pretraining Embedding in Fig. \ref{FIG:structure}), denoted as $E_p(\cdot)$, to extract a 512-dimensional feature vector and calculate the mean as the pre-trained class embedding of each class ($c_m=E_p(c)$). This strategy allows for more knowledge sharing from the peer dataset and improves the diversity of few-shot generation. This method has also been investigated recently by \cite{casanova2021instance,sauer2022stylegan}, but we adopt it specifically to share more knowledge from the peer dataset for few-shot image generation.

\noindent\textbf{Separable Class Modulation.} 
Original conditional GANs \cite{brock2018large,karras2020analyzing} inject conditional information at each resolution of the generated model to ensure the strength of the class conditions, which is unsuitable for PIP. We aim to strike a balance between class strength and information sharing, and therefore propose a separable class modulation method. Specifically, we use two separate mapping networks, $f_w(\cdot)$ and $f_c(\cdot)$, to map $z$ and $c$ to $\mathcal{W}$ and $\mathcal{C}$ spaces, respectively. The class mapping network $f_c$ has fewer layers than the $f_w$ network due to the limited amount of class information. This process can be formalized as:

\begin{equation}
	\begin{aligned}
	    w=f_w(z)\in\mathcal{W},\quad \hat{c}_m=f_c(c_m)\in\mathcal{C}.
	\end{aligned}
\end{equation}

\begin{table*}[t!]
\centering
\caption{FID metric (lower is better) and intra-LPIPS metric (higher is better) over Sunglasses, Sketches, Obama datasets. Apart from MixDL (LFS), StyleGAN2-ADA (LFS), and ours (PIP), other methods (TL) all pretrain on FFHQ and finetune to target datasets. Results with $^*$ are obtained when we run the official code of reference. Following \cite{ojha2021few}, intra-LPIPS is computed with 1K generated images. 
\small	\label{Tab:fid_free}}
\begin{tabular}{c|c|cc|cc|cc} 
\hline
&\multirow{2}{*}{Methods}&\multicolumn{2}{c|}{FFHQ$\rightarrow$Sunglasses}&\multicolumn{2}{c|}{FFHQ$\rightarrow$Sketches}&\multicolumn{2}{c}{FFHQ$\rightarrow$Obama}\\
&& FID ($\downarrow$) &  intra-LPIPS ($\uparrow$) &FID ($\downarrow$) &  intra-LPIPS ($\uparrow$)&FID ($\downarrow$) &  intra-LPIPS ($\uparrow$) \\
\hline 
	\multirow{9}{*}{TL}&TGAN \cite{wang2018transferring} & $55.61 $ &-& $53.42$&0.394&-&- \\
&TGAN+ADA \cite{karras2020training}  & $53.64$ &-& $66.99$ &0.414&-&-\\
&BSA \cite{noguchi2019image} & $76.12 $&- & $69.32 $&-&-&- \\
&FreezeD \cite{mo2020freeze} & $51.29 $ &$0.337^*$& $46.54 $&0.353&$62.4^*$&$0.19^*$\\
&MineGAN \cite{Wang_2020_CVPR} & $68.91 $&$0.402^*$ & $64.34 $ & 0.405&$75.6^*$&$0.23^*$\\
&EWC \cite{NEURIPS2020_b6d767d2} & $59.73 $ &$0.411^*$& $71.25 $ &0.421&$71.8^*$&$0.24^*$\\
&CDC \cite{ojha2021few} & $42.13 $ & $0.562^*$&$45.67 $ &0.453&$67.2^*$&$0.35^*$\\
&DCL \cite{zhao2022closer} & ${3 8 . 0 1} $&- & ${3 7 . 9 0} $&0.486&-&-\\
& RSSA \cite{xiao2022few}&$77.77 ^*$&$0.563^*$&$70.41 ^*$&\bm{$0.534^*$}&$88.86^*$&$0.36^*$\\
&ADAM \cite{zhao2022few}&$32.17^*$&$0.592^*$&$55.74$&$0.494^*$&$76.99^*$&$0.56^*$\\
\hline
\multirow{2}{*}{LFS}&StyleGAN2-ADA \cite{karras2020training}&$71.43^*$&$0.245^*$&$48.20^*$&$0.220^*$&{$57.39^*$}&$0.26^*$\\
& MixDL \cite{kong2021smoothing}&$61.97 ^*$&$0.414^*$&$55.40 ^*$&$0.400^*$ &$68.8^*$&$0.53^*$\\
\hline
PIP& Ours &\textbf{29.28}&\textbf{0.668}&\textbf{37.40 }&0.440&\textbf{55.52}&$\textbf{0.59}$\\

\hline
\end{tabular}
\end{table*}
Accordingly, we transform $w$ and $\hat{c}_m$ into style code and class code using learned affine transformations $A_s(\cdot)$ and $A_c(\cdot)$, respectively. The style code is then inserted into the synthesis network $g (\cdot)$ through the style modulation component (s-Mod) at each convolution layer, while the class code is controlled by a Key function $K(\cdot)$. Only a part of the class code is inserted into the synthesis network $g (\cdot)$ through the class modulation component (c-Mod), which improves knowledge sharing from the peer dataset effectively. This can be formalized as follows:
\begin{equation}
	\begin{aligned}
	    \text{style code}&:= A_s(w), \quad \text{class code}:= K(A_c(\hat{c}_m)),\\
	    \text{s-Mod}&:=s_i\cdot w_{i,j,k}, \quad \text{c-Mod}:=c_i\cdot w_{i,j,k},
	\end{aligned}
\end{equation}
where $w_{i,j,k}$ is original weight of the Synthesis network. $s_i$ and $c_i$ are the style scale and class scale corresponding to the $i$-th input feature map, respectively, and $j$ and $k$ enumerate the output feature maps and spatial footprint of the convolution, respectively. $K(\cdot)$ is a Key function that only allows part of the class code to pass and participate in the c-Mod. Accordingly, the whole weight demodulation of StyleGAN2 has been replaced in this paper as:

\begin{equation}
\begin{aligned}
w_{i j k}^{\prime \prime}&=w_{i j k}^{\prime} \bigg/ \sqrt{\sum_{i, k} w_{i j k}^{\prime}{ }^2+\epsilon},\
w_{i j k}^{\prime}&=s_i\cdot c_i \cdot w_{i,j,k}.
\end{aligned}
\end{equation}

Here,$\epsilon$ is a small constant to avoid numerical issues. $w_{i j k}^{\prime}$ and $w_{i j k}^{\prime \prime}$ are the modulated and demodulated weights, respectively. Separating style and class modulations was also recently investigated by \cite{laria2022transferring}. In contrast to our method, \cite{laria2022transferring} adopts a hyper-modulation to transfer a pre-trained unconditional GAN to a conditional GAN and does not have the Key component to reduce the class injection.

\subsection{Adapting Regularization}
\label{section_3.3}
Regularization is crucial for the training of GANs \cite{tseng2021regularizing,yang2021data,li2022fakeclr}. In this section, we introduce a direction regularization to improve the diversity in few-shot image generation.

\noindent\textbf{Direction Regularization.} To further enhance the diversity of few-shot generation, we adopt direction regularization using a pre-trained CLIP \cite{radford2021learning} model. Recent works \cite{zhang2022towards,gal2021stylegan,kwon2022one,zhu2021mind} demonstrate that CLIP-based methods can provide accurate domain characteristics in TL. In PIP, given the same latent code $z$, peer label $c_1$, and target label $c_2$, we aim for the sample-shift direction $\Delta d_{\text{sample}}$ to be parallel to the domain direction $\Delta d_{\text{domain}}$. The direction loss is given by:

\begin{equation}
	\begin{aligned}
	    \Delta d_{\text{sample}}&=E_I\left(G(z,c_1))-E_I(G(z,c_2)\right),\\
	    \Delta d_{\text{domain}}&=E_T(t_\text{peer})-E_T(t_\text{target}),\\
	    \mathcal{L}_\text{direction}&=1- \frac{\Delta d_{\text{sample}}\cdot\Delta d_{\text{domain}}}{|\Delta d_{\text{sample}}||\Delta d_{\text{domain}}|},
	\end{aligned}
\end{equation}
where $E_I$ and $E_T$ refer to pre-trained CLIP image and text encoders, respectively, while $t_\text{peer}$ and $t_\text{target}$ represent the text labels of the peer dataset and target dataset. Our direction loss aligns the target-generated space with the diverse peer-generated space, which greatly improves the diversity of the few-shot image generation. Unlike TL, which fine-tunes the pre-trained generator, PIP trains the generator from scratch. To prevent excessive direction loss from leading to cGAN training divergence and degradation, we employ lazy regularization, where direction loss is performed only once every 16 mini-batches.

In addition, we adopt regularization removal in our model. Path length regularization and style mixing are regularization techniques used in unconditional StyleGAN2 on single-class datasets like FFHQ to improve the continuity and editability of the latent space. However, these regularization techniques may be unnecessary and even detrimental on multi-modal datasets \cite{karras2021alias,sauer2022stylegan}. In our PIP setting, we have found that these regularization technologies are not suitable, and we have therefore removed path length regularization and style mixing from the original StyleGAN2.

\begin{figure*}[t!]
\setlength{\abovecaptionskip}{0.1cm}
    \setlength{\belowcaptionskip}{-0.3cm}
	\centering
	\includegraphics[scale=0.75]{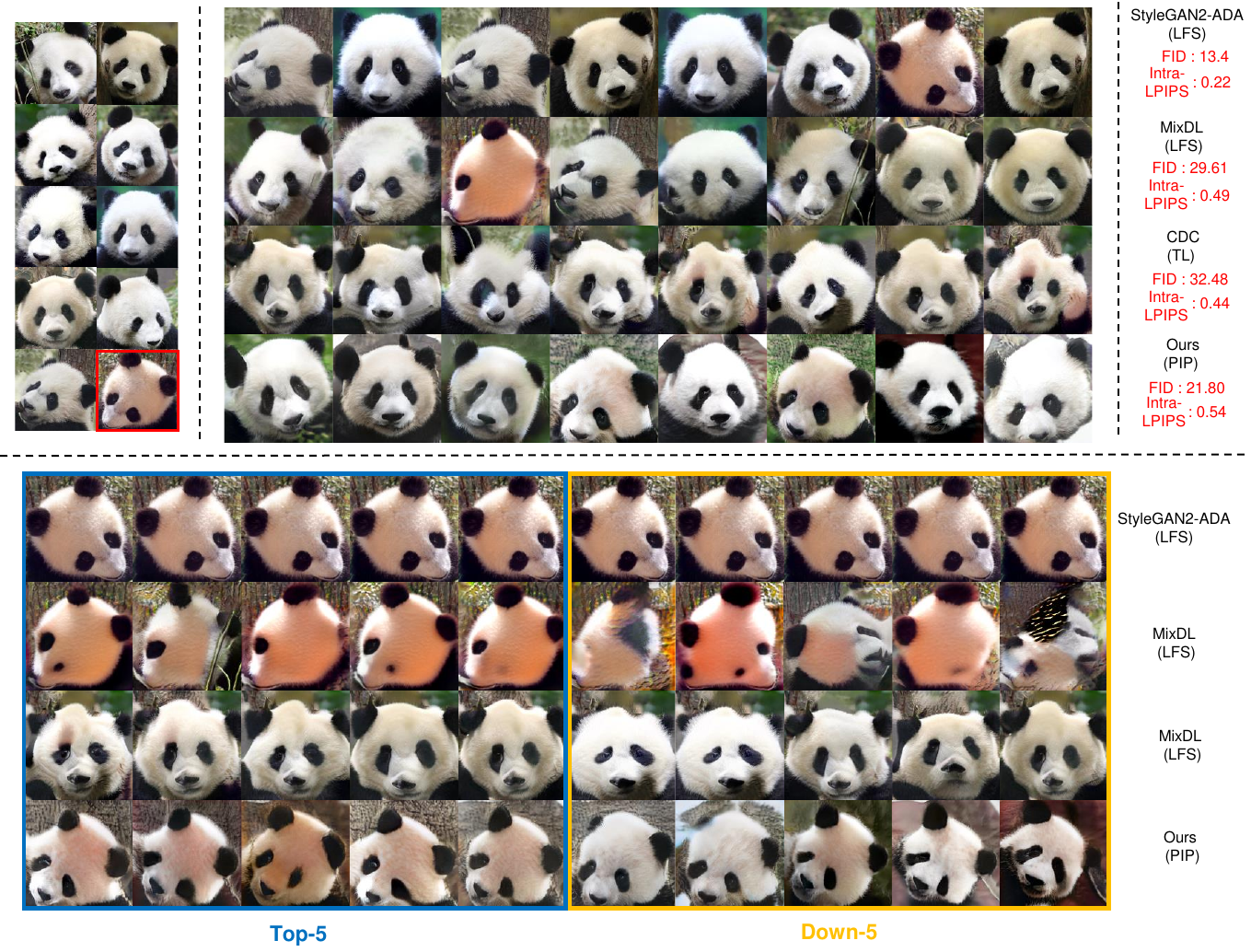}
	\caption{\textbf{Top-Left:}The 10-shot Panda dataset used as the target dataset. We compare the results of three pipelines and our method in terms of qualitative (\textbf{Top-Middle}) and quantitative evaluation (\textbf{Top-Right}).  \textbf{Bottom:} To evaluate the visual similarity of the generated images, we generated a large number of images (size=4,000) and assigned them to one of the 10 target samples (with the lowest standard LPIPS), resulting in 10 clusters. We then chose the cluster containing the \textcolor[RGB]{255,0,0}{reference image} and computed the LPIPS distance between the images within that cluster and the reference image. The \textcolor[RGB]{0,112,192}{Top-5 similar} and \textcolor[RGB]{255,192,0}{Bottom-5 similar} images are those with the five lowest and highest LPIPS distances, respectively. The images generated by LFS closely replicated the reference image, while the images generated by CDC (TL) differed significantly from the reference image. This indicates that CDC does not perform well when the source dataset has a large semantic gap with the few-shot dataset.}
	\label{FIG:panda}
\end{figure*}

\begin{figure*}[htp]
\setlength{\abovecaptionskip}{0.1cm}
	\begin{center}
		\includegraphics[width=1\linewidth]{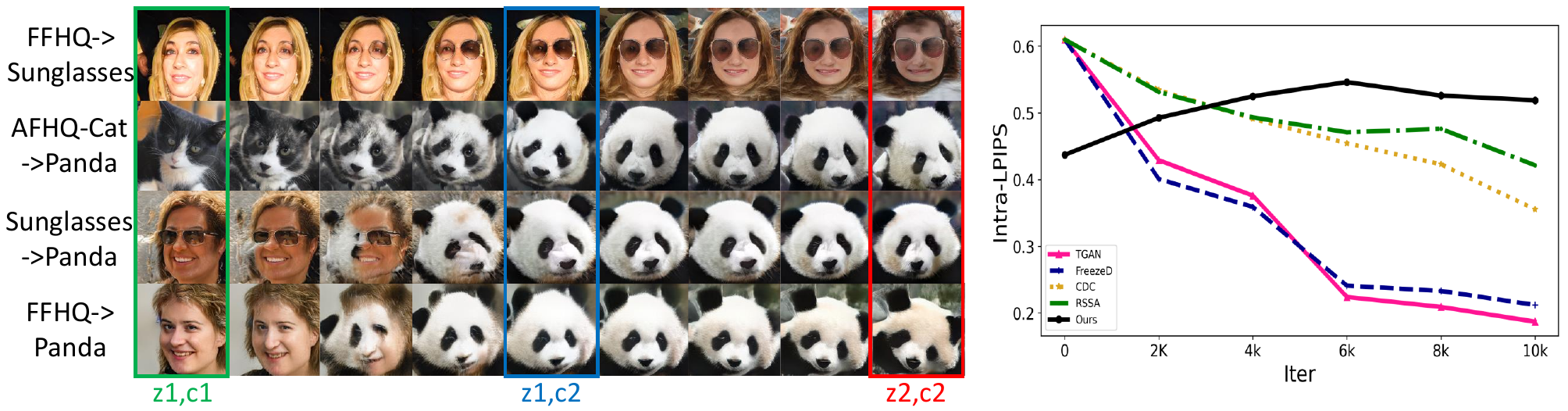}
	\end{center}
	\caption{\textbf{Left Part showcases the latent interpolation of the class space (between the \textcolor[RGB]{0,176,80}{peer class (z1, c1)} and \textcolor[RGB]{0,112,192}{target class (z1, c2)}) and the latent space (between the \textcolor[RGB]{0,112,192}{target class (z1, c2)} and \textcolor[RGB]{255,0,0}{(z2, c2)}). The amount of knowledge shared between peer and target classes depends on their domain gap. For instance, when the domain gap is narrow (e.g., FFHQ $\rightarrow$ Sunglasses), a significant amount of information, including pose, hairstyle, and facial features, is shared. When the domain gap is moderate (AFHQ-Cat $\rightarrow$ Panda), global knowledge such as pose is shared. However, when the domain gap is large, the interpolation exhibits discontinuity, and only limited information is shared. Furthermore, the generated samples in the target maintain continuity in their latent space across all settings. Right part showcases the Intra-LPIPS with training iteration on AFHQ-Cat$\rightarrow$ Panda dataset. TL methods exhibit high diversity in the early stages of the training process. However, as the training progresses, a gradual decline in diversity is observed.}}
	\label{fig:i-lpips}
\end{figure*}

\section{Experiments}
\label{eperiments}

In this section, we present the results of our experiments, which involve a comparison on datasets owning semantically related large source dataset (discussed in Sec. \ref{Free dataset}) and fixing the target dataset (10-shot Panda) and vary the peer datasets from a close to large domain gap (from AFHQ-Cat, Sunglasses, to FFHQ) (discussed in Sec. \ref{Restricted dataset}), as well as ablation studies (discussed in Sec. \ref{ablation}).

\noindent\textbf{Implementation.} We adopt the StyleGAN2 \cite{karras2020analyzing} as the backbone for our experiments, as it is commonly used in few-shot image generation tasks. Additionally, we use the adaptive data augmentation in StyleGAN2-ADA \cite{karras2020training} to alleviate the discriminator overfitting. We use the official implementation of \href{https://github.com/NVlabs/stylegan2-ada}{StyleGAN2-ADA} \cite{karras2020training} for our experiments, with the other training parameters and settings set to their default values.

\noindent\textbf{Datasets.} In this section, we present the datasets used for evaluation. In Sec. \ref{Free dataset}, we evaluate our method on Sketch, FFHQ-Sunglasses, and Obama datasets, which consist of approximately 300, 2500, and 100 images, respectively. To simulate few-shot generation setting, we choose 10-shot images as the training dataset and evaluate the performance of fitting the true distribution with the entire dataset. For previous transfer learning (TL) methods, we follow their settings and adopt the model pre-trained on the FFHQ dataset (which contains 70K images) as the source model and 10-shot images as the target dataset. In our PIP method, we randomly sample 1000 or 500 images from the FFHQ dataset \cite{karras2019style} as the peer dataset, and 10-shot images as the target dataset.
In Sec. \ref{Restricted dataset}, we fix the target dataset to 10-shot Panda and vary the peer datasets from close to large domain gaps (from AFHQ-Cat, Sunglasses, to FFHQ). Similar to Sec. \ref{Free dataset}, the source datasets (AFHQ-Cat, Sunglasses, and FFHQ) in TL consist of approximately 5K, 2.5K, and 70K images, respectively. In PIP, we choose 1K images from the source dataset to conduct the peer dataset. In the Sec. \ref{sec:exp_other}, we further evaluate our method on the Grumpy Cat dataset (using AFHQ-Cat as the peer dataset) and the Anime-face dataset (using FFHQ as the peer dataset). Both two datasets contain 100 images, and we choose 10-shot images as the training dataset to simulate few-shot generation setting and evaluate the performance of fitting the true distribution with the entire dataset. All images in our paper are resized to 256×256.

\noindent\textbf{Evaluation Metrics.} We adopt Fréchet Inception Distance (FID) \cite{heusel2017gans} and Intra-Cluster LPIPS (intra-LPIPS) \cite{ojha2021few,zhao2022closer} as the primary evaluation metrics. FID is the most widely used metric for image generation, and we calculate it between 5,000 generated images and the entire target dataset. Additionally, we use intra-LPIPS to measure the synthesis diversity. Specifically, we synthesize 1,000 images and assign each synthesized image to the $k$ training images with the lowest LPIPS distance, forming $k$ clusters. We then calculate the average LPIPS distance within each cluster and average over all the clusters.


\begin{table*}[t!]
\caption{FID and intra-LPIPS metrics over experiments with different/large domain gap. Baseline result are obtained when we run the official code of reference. Our results demonstrate that PIP has greater robustness to increasing domain gaps, producing better results in terms of both reality and diversity compared to current state-of-the-art methods.
\label{Tab:fid_restrict}}
\centering

	\begin{tabular}{c|c|cc|cc|cc|cc}	
		\hline
        &Dataset (LPIPS-EMD)&\multicolumn{2}{c|}{FFHQ$\rightarrow$ Sunglass (0.68)}&\multicolumn{2}{c|}{Cat$\rightarrow$ Panda (0.71)}&\multicolumn{2}{c|}{Sunglass$\rightarrow$ Panda (0.75)}&\multicolumn{2}{c}{FFHQ$\rightarrow$ Panda (0.77)}\\
	&	Methods& FID ($\downarrow$) & i-LPIPS ($\uparrow$)& FID & i-LPIPS& FID & i-LPIPS&FID & i-LPIPS\\
\hline 
\multirow{5}{*}{TL}&MineGAN \cite{Wang_2020_CVPR}&68.91&0.40&67.29&0.22&70.53&0.24&69.81&0.23\\
&EWC \cite{NEURIPS2020_b6d767d2}&59.73&0.41&53.71&0.23&66.68&0.23&63.27&0.25\\
&CDC \cite{ojha2021few}&42.13&0.56&32.48&0.44&35.56&0.40&31.40&0.50\\
&RSSA \cite{xiao2022few}&77.77&0.56&135.0&0.52&142.9&0.42&128.8&0.49\\
&ADAM \cite{zhao2022few}&32.17&0.59&239.6&0.52&205.7&0.49&215.9&0.54\\
\hline
PIP&Ours&\textbf{27.83}&\textbf{0.67}&\textbf{15.67}&\textbf{0.52}&\textbf{21.77}&\textbf{0.50}&\textbf{22.13}&\textbf{0.54}\\
		\hline
	\end{tabular}
\end{table*}

\begin{table*}[t!]
\caption{Ablation study on different components of our method.
	\label{Tab:ablation_1}}
\centering
\begin{tabular}{l|cc|cc|cc}
\hline 
Dataset&\multicolumn{2}{c}{Sunglasses}&\multicolumn{2}{c}{Sketches}&\multicolumn{2}{c}{Obama} \\
\hline
Configuration & FID $\downarrow$ & intra-LPIPS $\uparrow$&  FID & intra-LPIPS& FID & intra-LPIPS\\
\hline 
$\mathbf{A} $  StyleGAN2-ADA (U) &71.43 &0.25&48.20&0.22&57.39 &0.26\\
$\mathbf{B} $  StyleGAN2-ADA (C) &  44.12&0.56&39.84&0.37&73.77&0.50\\
$\mathbf{C}$  $+$ Separable class modulation &  34.51&0.62&48.17&0.39&68.69&0.52\\
$\mathbf{D} $  $+$ Pretrained class embedding & 32.53&0.62&46.12&0.41&58.38&0.53 \\
$\mathbf{E}$  $+$ Regularization removing &  31.83&0.64&41.28&0.40&60.19&0.54\\
$\mathbf{F} $  $+$ Direction regularization &  29.28&0.67&37.40&0.44&55.52&0.59 \\
\hline
\end{tabular}
\end{table*}

\begin{figure*}[htp]
\setlength{\abovecaptionskip}{0.1cm}
    \setlength{\belowcaptionskip}{-0.3cm}
	\centering
	\includegraphics[scale=0.75]{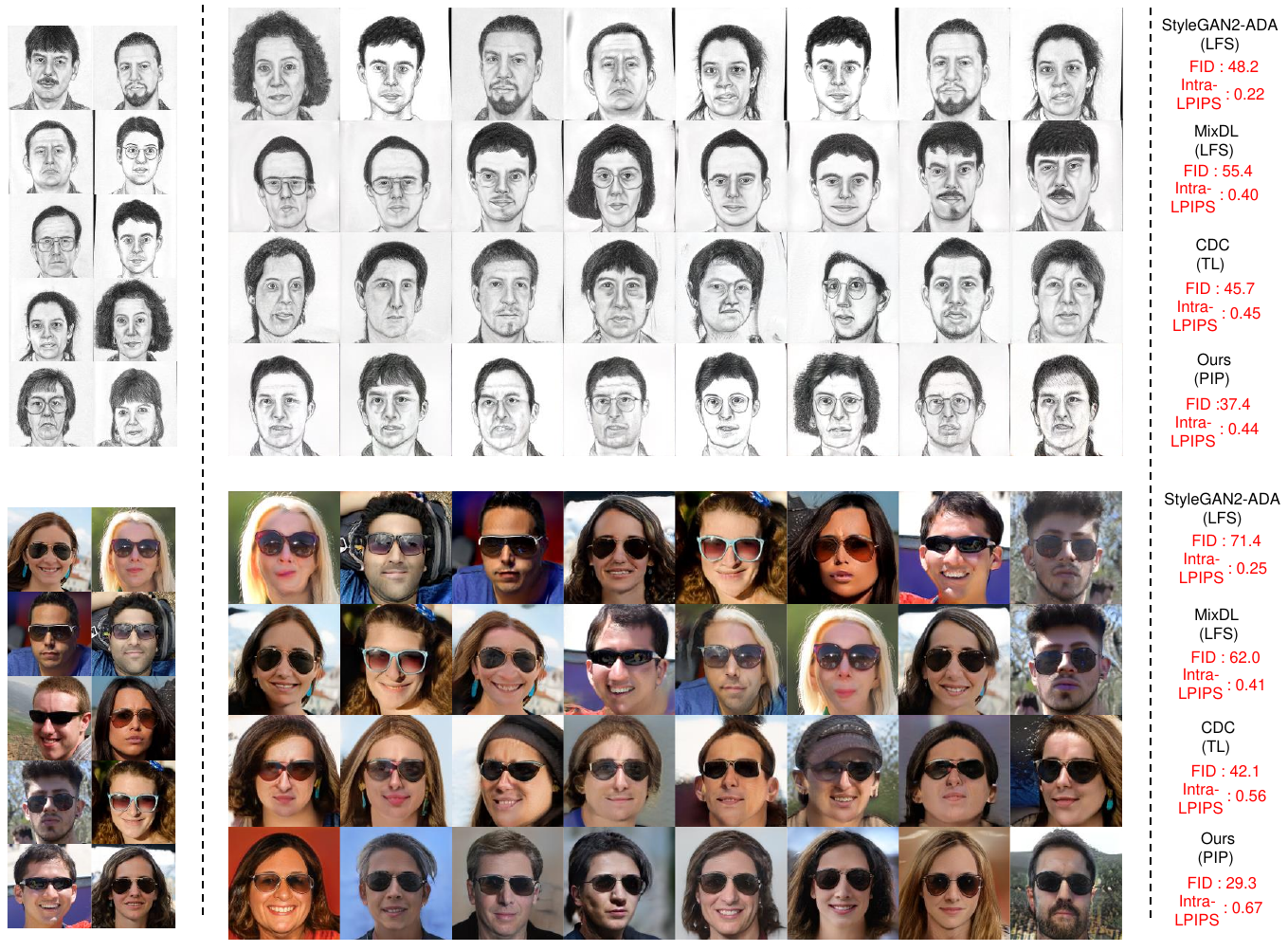}
	\caption{\textbf{Left:} Target samples on 10-shot Sketch and Sunglasses datasets. \textbf{Mid:} We showcase the generated results of three pipelines. StyleGAN 2-ADA (LFS) tends to replicate the training data, while MixDL (LFS) can be used as a regularization method to mitigate this issue slightly. CDC (TL) is pretrained on a large-scale source dataset, which provides enough diversity and realism. Our method (PIP) combines 1k images from a peer dataset and the 10-shot target dataset, effectively acquiring shared knowledge from the peer dataset. This approach meets all three desired properties of few-shot image generation \textbf{Right:} Metrics (FID and Intra-LPIPS) of different methods.}
	\label{FIG:result}
\end{figure*}

\begin{figure*}[htp]
	\centering
	\includegraphics[scale=0.75]{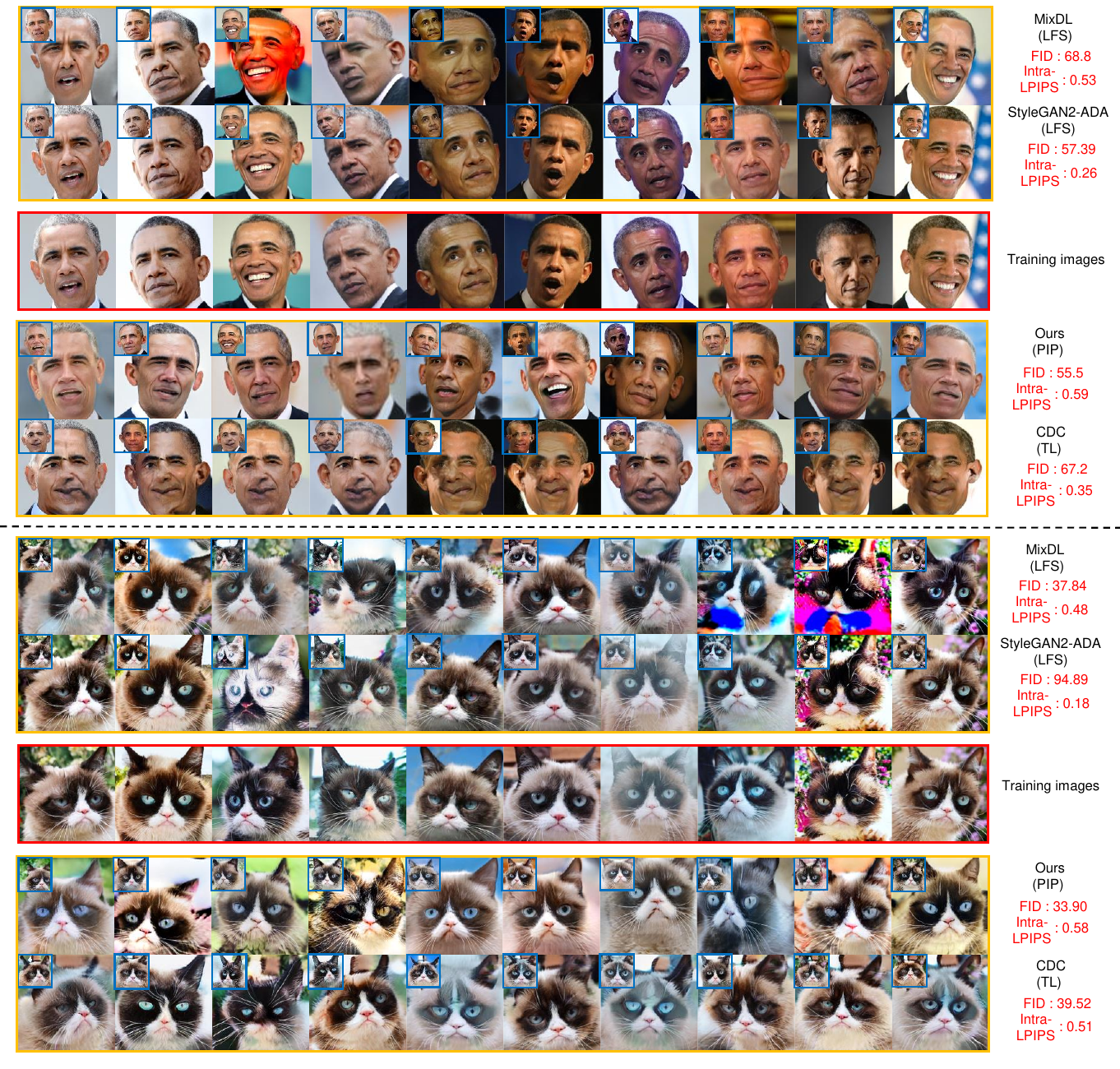}
	\caption{Generated images on Obama and Grumpy Cat datasets. {Middle line} is the \textcolor[RGB]{255,0,0}{reference image} on 10-shot training dataset. The other lines show the generated images and corresponding metrics (FID and Intra-LPIPS) of both baselines and our method. The big images are generated images \textcolor[RGB]{255,192,0}{Down similar} to reference image on training images and the small images are generated images \textcolor[RGB]{0,112,192}{Top similar} to reference image on training images. Specifically, we generated a large number of images (size=4,000) and assigned them to one of the 10 target samples (with the lowest standard LPIPS), resulting in 10 clusters. We then chose the cluster containing the \textcolor[RGB]{255,0,0}{reference image} and computed the LPIPS distance between the images within that cluster and the reference image. The \textcolor[RGB]{0,112,192}{Top similar} and \textcolor[RGB]{255,192,0}{Bottom similar} images are those with the lowest and highest LPIPS distances, respectively.}
	\label{FIG:experiment_1}
\end{figure*}

\subsection{Results on Datasets Owning Semantically Related Large Source Dataset}
\label{Free dataset}
In experiments on datasets with semantically related large source dataset, we select three common datasets, Sunglasses, Sketches, and Obama. 
Table \ref{Tab:fid_free} presents a comparison of FID and intra-LPIPS with previous studies. We compared our results with the most recent TL and LFS methods. LFS methods, such as \cite{karras2020training, kong2021smoothing}, showed poor performance on all datasets, as measured by both FID and intra-LPIPS. On the other hand, TL methods showed significant improvement on these datasets, especially for intra-LPIPS on the Sketches dataset, with the use of semantically related large source datasets. Notably, our proposed method, which only uses 1K peer data, achieves remarkable results on most datasets. Furthermore, we present a qualitative comparison of our PIP method with a popular baseline on three tasks: FFHQ $ \rightarrow $ Sketch (top part of Fig. \ref{FIG:result}), FFHQ $ \rightarrow $ Sunglasses (bottom part of Fig. \ref{FIG:result}), and FFHQ $ \rightarrow $ Obama (top part of Fig. \ref{FIG:experiment_1} ). Our PIP method has demonstrated significant improvement in terms of both realism and diversity.


\subsection{Experiments with different/large domain gap}
\label{Restricted dataset}

\begin{table*}[t!]
\centering
\caption{Ablation study on size of peer dataset with different domain gap. \label{Tab:ablation_2}}
	\centering
  \setlength{\tabcolsep}{1.0mm}{
	\begin{tabular}{c|cc|cc|cc|cc}	
		\hline
        Dataset (LPIPS-EMD)&\multicolumn{2}{c|}{FFHQ $\rightarrow$ Sunglass (0.68)}&\multicolumn{2}{c|}{Cat $\rightarrow$ Panda (0.71)}&\multicolumn{2}{c|}{Sunglass $\rightarrow$ Panda(0.75)}&\multicolumn{2}{c}{FFHQ $\rightarrow$ Panda (0.77)}\\
		Size of peer dataset& FID ($\downarrow$) & i-LPIPS ($\uparrow$)& FID & i-LPIPS& FID & i-LPIPS&FID & i-LPIPS\\
\hline 
200&50.49&0.59&26.16&0.48&35.36&0.45&110.29&0.31\\
500&41.99&0.60&21.23&0.51&24.42&0.48&27.28&0.50\\
1000&29.28&0.67&\textbf{15.67}&\textbf{0.52}&\textbf{21.77}&\textbf{0.50}&\textbf{22.13}&\textbf{0.54}\\
2000&\textbf{27.83}&\textbf{0.67}&88.19&0.49&92.34&0.45&104.31&0.40\\
		\hline
	\end{tabular}}
\end{table*}

\begin{table*}[t!]
\centering
\scriptsize
\caption{Ablation study on the key value of class modulation. LPIPS-EMD indicates the domain gap between peer and target domains. The size of peer dataset is set to 1000 in this experiment.\label{Tab:ablation_3}}
	\centering
	\begin{tabular}{l|cc|cc|cc|cc|cc}	
		\hline
        Dataset (LPIPS-EMD)&\multicolumn{2}{c|}{FFHQ$\rightarrow$ Obama (0.68)}&\multicolumn{2}{c|}{FFHQ$\rightarrow$ Sunglass (0.68)}&\multicolumn{2}{c|}{Cat$\rightarrow$ Panda (0.71)}&\multicolumn{2}{c|}{Sunglass$\rightarrow$ Panda (0.75)}&\multicolumn{2}{c}{FFHQ $\rightarrow$ Panda (0.77)}\\
		Key value & FID ($\downarrow$) & i-LPIPS ($\uparrow$)& FID & i-LPIPS& FID & i-LPIPS&FID & i-LPIPS&FID & i-LPIPS\\
\hline 
$\mathbf{A} :$  4& 73.28&0.53&{29.28}&{0.67}&27.19&0.55&33.61&0.51&34.58&0.56\\
$\mathbf{B} :$  8&76.89&0.55 &30.57&0.62&31.57&0.56&40.46&0.52&40.18&0.56\\
$\mathbf{D} :$  4+8&65.93&0.48 &32.60&0.61&22.90&0.54&28.90&0.51&30.52&0.54\\
$\mathbf{E} :$  4+16&71.77&0.57 &34.10&0.61&21.80&0.54&32.21&0.51&33.82&0.54\\
$\mathbf{F} :$  4+8+16 &70.31&0.53&31.45&0.62&19.89&0.53&26.19&0.51&27.44&0.54\\
$\mathbf{G} :$  4+8+16+64+128+256&55.52&0.59 &45.22&0.60&{15.67}&0.52&{21.77}&0.50&{22.13}&0.54\\
		\hline
	\end{tabular}
\end{table*}

In this section, we investigate the effectiveness of our proposed PIP method on the 10-shot Panda dataset with peer datasets that vary from close to large domain gaps (AFHQ-Cat, Sunglasses, to FFHQ). We use LPIPS-based Earth Movers Distance (EMD) as a basic measurement of the domain gap. EMD is a common distribution distance that has been used in many fields \cite{kolkin2019style}.  To calculate the EMD, we randomly
sample $m$ (1000) source images and $n$ (100) target images as the source domain set $I_S$ and the target domain set $I_T$, respectively. We extract the intermediate features of $I_S$ and $I_T$ from the image encoder $\mathcal{E}$ used in LPIPS. The LPIPS-based EMD is the mean LPIPS distance for each source feature from $F_S$ with its closest target feature from $F_T$, where $F_S={F^1_S,\cdots,F^m_S}$ and $F_T={F^1_T,\cdots,F^n_T}$ are the extracted intermediate features through $\mathcal{E}$. The LPIPS-EMD is defined as:
\begin{equation}
\label{LPIPS-EMD}
   \text{LPIPS-EMD} =\max \left(\frac{1}{n} \sum_i \min _j \boldsymbol{C}_{i, j}, \frac{1}{m} \sum_j \min _i \boldsymbol{C}_{i, j}\right),
\end{equation}
 where $\boldsymbol{C}$ is the LPIPS matrix to measure the sample-wise distances from $F_S$ to $F_T$, and each element is computed as:
$\boldsymbol{C}_{i, j}=\text{LPIPS}(F^i_S,F^j_T)$. Table \ref{Tab:fid_restrict} shows that PIP outperforms state-of-the-art TL methods consistently, although its performance decreases as the domain gap increases. We further provide qualitative comparisons and Intra-LPIPS with training iterations on the AFHQ-Cat$\rightarrow$Panda dataset in Figs. \ref{FIG:panda} and \ref{fig:i-lpips}, respectively, which demonstrate the effectiveness of our PIP pipeline and our method. Finally, we show the latent interpolations in class and latent spaces on datasets with different domain gaps in Fig. \ref{fig:i-lpips}. Our results indicate that the extent of knowledge sharing between peer and target classes is adaptively contingent upon their domain gap in our PIP, which is the main factor behind the advancement of our method.

 \begin{table}

\caption{FID and intra-LPIPS metrics over Grumpy Cat and Anime-face datasets. Baseline result are obtained when we run the official code of reference. Note that the Grumpy Cat and Anime-face datasets only contains 100 images, the large synthesis diversity will cause a great increase in the FID. Therefore, replicating the training images on StyleGAN2-ADA has the competitive FID results. However, the diversity measured by intras-LPIPS is poor for StyleGAN2-ADA on all datasets. 
	\label{Tab:extra_experments}}
 \scriptsize
\begin{tabular}{c|c|cc|cc} 
\hline
&\multirow{2}{*}{Methods}&\multicolumn{2}{c|}{Grumpy Cat}&\multicolumn{2}{c}{Anime-face}\\
 && FID ($\downarrow$) &  intra-LPIPS ($\downarrow$) & FID&intra-LPIPS \\
\hline 
	\multirow{2}{*}{LFS}& StyleGAN2-ADA \cite{karras2020training}&94.89&0.18&79.72&0.22\\
&MixDL \cite{kong2021smoothing}&37.54&0.48&82.34&0.46\\
\hline
\multirow{2}{*}{TL}& CDC \cite{ojha2021few}&39.52&0.51&95.24&0.48\\
& RSSA \cite{xiao2022few}&49.72&0.53&112.5&0.49\\
\hline
PIP& ours   &\textbf{33.90}&\textbf{0.58} &\textbf{78.88}& \textbf{0.52} \\
\hline
\end{tabular}
\end{table}

\subsection{Experiments on Grumpy Cat and Anime-face Datasets}
\label{sec:exp_other}
\begin{figure*}[t!]
\setlength{\abovecaptionskip}{0.1cm}
    \setlength{\belowcaptionskip}{-0.3cm}
	\centering
	\includegraphics[scale=0.75]{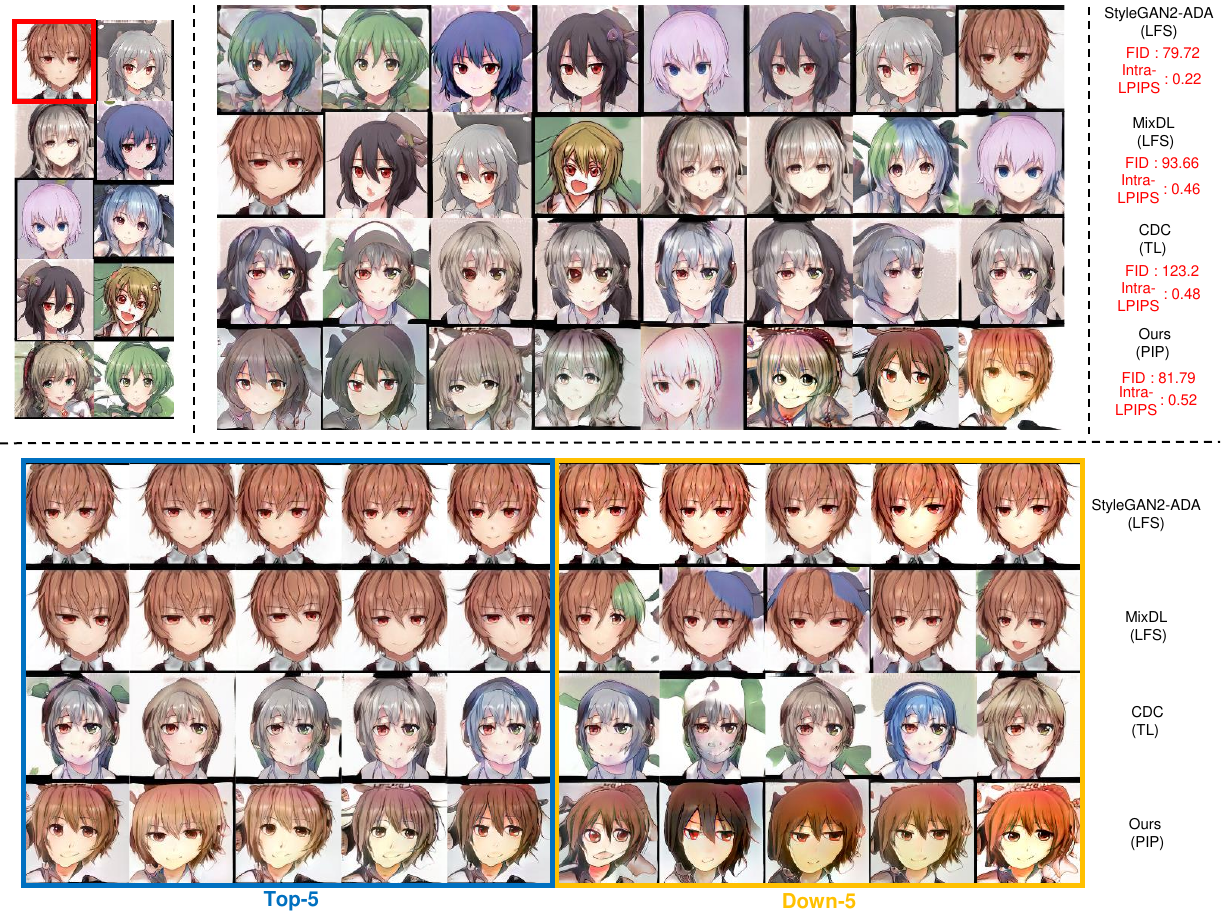}
	\caption{ \textbf{Top-Left:}The 10-shot Anime-face dataset used as the target dataset. We compare the results of three pipelines and our method in terms of qualitative (\textbf{Top-Middle}) and quantitative evaluation (\textbf{Top-Right}).  \textbf{Bottom:} To evaluate the visual similarity of the generated images, we generated a large number of images (size=4,000) and assigned them to one of the 10 target samples (with the lowest standard LPIPS), resulting in 10 clusters. We then chose the cluster containing the \textcolor[RGB]{255,0,0}{reference image} and computed the LPIPS distance between the images within that cluster and the reference image. The \textcolor[RGB]{0,112,192}{Top-5 similar} and \textcolor[RGB]{255,192,0}{Bottom-5 similar} images are those with the five lowest and highest LPIPS distances, respectively. The images generated by LFS closely replicated the reference image, while the images generated by CDC (TL) differed significantly from the reference image. This indicates that CDC does not perform well when the source dataset has a large semantic gap with the few-shot dataset.}
	\label{FIG:Anime-face}
\end{figure*}

In this section, we illustrate the quantitative (Tab \ref{Tab:extra_experments}) and qualitative comparison on Grumpt Cat (bottom part of Fig. \ref{FIG:experiment_1}) and Anime-face (Fig. \ref{FIG:Anime-face}) datasets. In this case, other methods have limited diversity and reality, and our method acquires significant improvement.

\subsection{Ablation Study}
\label{ablation}
\textbf{Effect of Different Components in Our Method.} Ablation studies conducted on the various components of our proposed method are presented in Table \ref{Tab:ablation_1}. The PIP pipeline (\textbf{Config-B}) was found to improve diversity but decrease reality when compared to the LFS pipeline (\textbf{Config-A}). However, the four technologies proposed in \textbf{Config-C, D, E, and F} significantly improved both reality and diversity in the generated images. Our method (\textbf{Config-F}) achieved an impressive FID improvement of $59\%$, $22.4\%$, and $3.3\%$ on the Sunglasses, Sketches, and Obama datasets, respectively, over StyleGAN2-ADA (U). Importantly, our method (\textbf{Config-F}) also achieved significant intra-LPIPS improvements on all datasets compared to the LFS method StyleGAN2-ADA (U).

\noindent\textbf{Effect of the Size of Peer Dataset.}
 When the peer dataset is excessively large, the PIP model faces an imbalanced conditional generation task between the peer and few-shot target datasets. This causes the model to focus excessively on generating the peer dataset, thereby compromising its ability to generate the few-shot target dataset. Consequently, a large peer dataset is unnecessary for our PIP method, which is a significant advantage over TL methods that require a substantial source dataset to pre-train the source generator, making it challenging to implement in real-world scenarios. In our PIP method, determining the appropriate size of the peer dataset is straightforward. As demonstrated in Table \ref{Tab:ablation_2}, excellent results can be achieved across all datasets by limiting the size of the peer dataset to 1K.
 



\noindent\textbf{Effect of the Key Value of Class Modulation.}
The function K($\cdot$) serves as a gate function that determines whether class modulation should be implemented. Key value equals '4' means that class injection is implemented only at the $4\times4$ feature layer of the generator. When the domain gap is large, more class injection is needed to improve performance. As shown in Table \ref{Tab:ablation_3}, a key value of '4+8+16+64+128+256' is used for Cat/Sunglasses/FFHQ $\rightarrow$ Panda datasets. Conversely, when the domain gap is narrow, a key value of '4' is sufficient, and using too much injection can reduce diversity and performance. Notably, transferring multiple identities to a specific person is challenging for both TL-based methods and our PIP, resulting in FFHQ$\rightarrow$Obama being an outlier in the monotonic trend. In summary, for settings with a narrow domain gap (FFHQ$\rightarrow$Sunglasses) or easy to complete (FFHQ$\rightarrow$Sketch), the key value is set to '4'. For settings with a moderate domain gap (FFHQ $\rightarrow$Anime-face and AFHQ Cat$\rightarrow$Grumpy Cat), the key value is set to '4+16'. For settings with a large domain gap (AFHQ-Cat/Sunglasses /FFHQ $\rightarrow$ Panda) or challenging to complete (FFHQ $\rightarrow$Obama), the key value is set to '4+8+16+64+128+256'.

\begin{table*}[h]
\scriptsize
\caption{Ablation on Pre-trained technologies. $*$ indicates that we set the class injection with resolution 4+8+16+64+128+256 instand of 4+16  on Panda generation task. The other settings are the same as original paper (Size of peer dataset is 1000).}
	\centering
 \setlength{\tabcolsep}{1.5mm}{
	\begin{tabular}{c|cc|cc|cc|cc|cc|cc|cc}	
		\hline
        Source dataset&\multicolumn{10}{c|}{FFHQ}&\multicolumn{4}{c}{AFHQ-Cat}\\
        \hline
        Target datset&\multicolumn{2}{c|}{Babies}&\multicolumn{2}{c|}{Sunglasses}&\multicolumn{2}{c|}{Sketches}&\multicolumn{2}{c|}{Obama}&\multicolumn{2}{c|}{Anime-face}&\multicolumn{2}{c|}{Grumpy Cat}&\multicolumn{2}{c}{$\text{Panda}^*$}\\
		Configuration & FID($\downarrow$) & i-LPIPS($\uparrow$)& FID & i-LPIPS& FID & i-LPIPS&FID & i-LPIPS&FID & i-LPIPS&FID & i-LPIPS&FID & i-LPIPS\\
        \hline 
        CDC&74.39&0.57&42.13&0.56&45.67&\textbf{0.45}&67.2&0.35&95.24&0.48&39.52&0.51&32.48&0.44\\
        RSSA&63.44&0.58&77.77&0.56&70.41&0.53&88.86&0.36&112.5&0.49&49.72&0.53&135.0&0.52\\
        \hline
        w/o VGG+w/o CLIP&72.75&0.62&34.19&0.63&43.32&0.42&64.67&0.56&84.29&0.49&35.02&0.55&28.32&0.48\\
        w/ VGG+w/o CLIP&66,32&0.60&31.83&0.64&41.28&0.40&60.19&0.54&82.31&0.51&34.27&0.54&20.38&0.49\\
        w/ VGG+w/ CLIP&\textbf{61.84}&\textbf{0.66}&\textbf{29.28}&\textbf{0.67}&\textbf{37.40}&0.44&\textbf{55.52}&\textbf{0.59}&\textbf{78.88}&\textbf{0.52}&\textbf{33.90}&\textbf{0.58}&\textbf{15.67}&\textbf{0.52}\\
		\hline
	\end{tabular}}
\label{tab:storage}
\end{table*}

\begin{table*}[t!]
\centering
\caption{Time and space cost of introducing the pre-trained model during training and inference. 
\label{Tab:time}}
\setlength{\tabcolsep}{1.0mm}{
	\begin{tabular}{c|cc|cc}	
		\hline
       Backbones&\multicolumn{2}{c|}{Training}&\multicolumn{2}{c}{Inference}\\
	& w/ CLIP\&VGG & w/o CLIP\&VGG & w/ CLIP\&VGG & w/o CLIP\&VGG\\
\hline 
Time cost (Second/kimg)&11.05& 10.53 & 6.55& 6.54\\
Space cost (G) &32 & 25&17&16\\
\hline
\end{tabular}}
\end{table*}
\noindent\textbf{Effect of the Pre-trained class embedding and CLIP.}
Our proposed Progressive Image Processing (PIP) approach necessitates the utilization of pre-trained class embedding and Contrastive Language–Image Pretraining (CLIP) models. This characteristic sets it apart from conventional train-from-scratch methodologies. Moreover, the prerequisite for pre-trained models within our PIP framework is less intricate to satisfy compared to traditional Transfer Learning (TL) techniques. Initially, the pre-trained CLIP technique finds widespread employment in TL methodologies, as evidenced by its application in recent works such as \cite{zhang2022towards,zhang2022generalized}. Furthermore, the VGG model adopted in our study has undergone pre-training on the ImageNet dataset, while CLIP has been pre-trained on a collection of image-text pairs. This pre-training renders both models versatile and amenable to deployment across diverse datasets. Notably, both the CLIP and VGG models are conveniently accessible online and obviate the need for supplementary training efforts.

In order to comprehensively assess the impact of pre-trained VGG and CLIP models on our PIP approach, we conducted an ablation study involving various pre-training techniques, the results of which are detailed in Table \ref{tab:storage}. It is imperative to acknowledge that although the integration of pre-trained CLIP and VGG models demonstrably enhances the performance of our approach, the methodology lacking pre-trained class embedding and CLIP regularization still yields noteworthy improvements over conventional TL methodologies.

Moreover, we emphasize that the incorporation of pre-trained VGG and CLIP encoders within our approach yields compelling cost-effectiveness. The CLIP embedding is exclusively indispensable during the training phase and exerts negligible adverse ramifications on time and memory consumption during inference. Simultaneously, the VGG network's lightweight nature enables seamless forward propagation, incurring minimal temporal costs. To substantiate our assertions, we provide a comprehensive breakdown of the temporal and spatial costs incurred by the integration of pre-trained models during both the training and inference phases, as documented in Table \ref{Tab:time}. All experimental trials were executed on a hardware setup comprising 2 A100 GPUs.

\section{Limitation and Conclusion}
\noindent\textbf{Limitation.}
Although our proposed method has achieved remarkable generation results on datasets without semantically related large source dataset, it has not outperformed state-of-the-art methods with TL pipeline in terms of reality and diversity on some datasets with semantically related large source dataset. For example, as shown in Table \ref{Tab:fid_free}, our method shows poorer diversity compared to state-of-the-art methods on the Sketch dataset. Another limitation of our method is that it relies on layer-wise information injection in the generator, where the latent code is fed into all convolution layers. Fortunately, this layer-wise structure is commonly used in state-of-the-art GANs such as \cite{brock2018large,karras2019style,karras2020analyzing,sauer2022stylegan}. Additionally, our Key value is manually set through experience and may not be suitable for all datasets. An adaptive method may further improve the performance.


\noindent\textbf{Conclusion.}
Our paper proposes a novel few-shot image generation pipeline called Peer is your Pillar (PIP), which combines a target few-shot dataset with a peer dataset to create a data-unbalanced conditional generation. To enhance the diversity of the generated images, we also present several new techniques including Pre-trained Class Embedding, Separable Class Modulation, and Direction Regularization. Our approach is uniquely capable of adapting to different knowledge preservation levels from the source model, making it ideal for scenarios where the source and target domains have varying degrees of relatedness.  Through qualitative and quantitative evaluations, we demonstrate that our approach significantly outperforms previous methods and achieves all three desired properties in few-shot image generation.

\section*{Acknowledgments}
The work was supported in part by the National Natural Science Foundation of China under Grands U19B2044 and
61836011


{\small
\bibliographystyle{ieee_fullname}
\bibliography{PIP.bib}
}

\vfill

\end{document}